# Pre-Deployment Testing of Low Speed, Urban Road Autonomous Driving in a Simulated Environment

Xinchen Li, Aravind Chandradoss Arul Doss, Bilin Aksun Guvenc, and Levent Guvenc

Automated Driving Lab, Ohio State University

## Abstract

Low speed autonomous shuttles emulating SAE Level L4 automated driving using human driver assisted autonomy have been operating in geo-fenced areas in several cities in the US and the rest of the world. These autonomous vehicles (AV) are operated by small to mid-sized technology companies that do not have the resources of automotive OEMs for carrying out exhaustive, comprehensive testing of their AV technology solutions before public road deployment. Due to the low speed of operation and hence not operating on roads containing highways, the base vehicles of these AV shuttles are not required to go through rigorous certification tests. The way these vehicles' driver assisted AV technology is tested and allowed for public road deployment is continuously evolving but is not standardized and shows differences between the different states where these vehicles operate. Currently, AVs and AV shuttles deployed on public roads are using these deployments for testing and improving their technology. However, this is not the right approach. Safe and extensive testing in a lab and controlled test environment including Model-in-the-Loop (MiL), Hardware-in-the-Loop (HiL) and Autonomous-Vehicle-in-the-Loop (AViL) testing should be the prerequisite to such public road deployments. This paper presents three dimensional virtual modeling of an AV shuttle deployment site and simulation testing in this virtual environment. We have two deployment sites in Columbus of these AV shuttles through the Department of Transportation funded Smart City Challenge project named Smart Columbus. The Linden residential area AV shuttle deployment site of Smart Columbus is used as the specific example for illustrating the AV testing method proposed in this paper.

## Introduction

Several cities in the U.S. are working on transforming into smart cities with smart mobility choices for their residents. These smart mobility choices include the use of Autonomous Vehicles (AV) and smart road infrastructure enabling connected vehicle applications. The overall smart mobility aim is to make it convenient and safer for residents or passengers to commute or travel on designated routes within the city. AV shuttles operating on fixed routes in geo-fenced areas are being used as one of these smart mobility choices in smart cities. These are low speed shuttles with assisted autonomy that emulate SAE level L4 automated driving and are operated by start-up companies or relatively small technology companies and not automotive Original Equipment Manufacturers (OEM). Since these AV shuttles are deployed on public roads in urban areas with other low speed traffic and pedestrians and bicyclists without being subjected to a rigorous testing and certification procedure, safety and its assessment becomes a very important concern. According to recent data for the U.S. from NHTSA [1], an estimated 36,750 people were killed in traffic accidents and car crashes in the year 2018 in the U.S. The introduction of AVs and AV shuttles into public road usage should not increase that number but should rather decrease it by avoiding accidents from happening. Unified, scalable and replicable approaches like what was proposed in our earlier work reported in [2] can also be used as a basis for a unified approach for evaluation of these AVs to assess their readiness and safety for public road deployment.

Due to the low speed of operation and hence not operating on roads containing highways, the base vehicles of these AV shuttles are currently not required to go through rigorous certification tests. The way these vehicles' driver assisted AV technology is tested and allowed for public road deployment is continuously evolving but is not standardized and shows differences between the different states where these vehicles operate. Safety of operation is achieved by first limiting the speed of operation to be below 25 mph, making sure that the driver takes over during difficult maneuvers, if necessary, like handling intersections and using simple collision avoidance to stop automatically or through driver intervention every time an obstacle is encountered on the pre-determined path. Past experience with AV deployments in public roads has shown that while this low speed human assisted autonomy is a safe mode of operation, it is not capable of preventing accidents altogether as evidenced by several cases to the contrary including: un-attentive drivers not being able to take over control fast enough. Some examples to support this are AVs being rear ended by other vehicles due to their driving pattern being unexpected for other road users, sudden emergency braking of the AV resulting in passengers being hurt, not being able to backup manually as an un-attentive truck backing into the road is definitely going to cause a collision and, yes, a pedestrian hitting the side of the AV shuttle. Those and many more unexpected and somewhat rare traffic scenarios must be tested before deploying an AV shuttle on public roads.

While the larger technology companies and OEMs have the resources to do more exhaustive testing, it is more difficult for individual startup companies and AV technology researchers to develop such tools independently. What is, therefore, proposed and presented for the case of AV deployments in Smart Columbus in this paper is the use of open source and publicly shared and realistic simulation environments where such tests can take place in Model-in-the-Loop (MiL) and Hardware-in-the-Loop (HiL) testing. This paper presents such an approach to AV simulation evaluation, using the AV shuttle deployment in Columbus, Ohio as specific example for illustration of the method. This paper, thus, explains first, how the Linden Residential Area AV shuttle deployment site considered in Columbus, Ohio was modeled using two different approaches. The rest of the paper concentrates on the Linden AV shuttle route and illustrates how the three dimensional (3D) computer representation of the environment is used to build a 3D point cloud based map, later combined with a vector map. To illustrate the potential use of such a simulator environment, an AV shuttle with a soft 3D lidar sensor is operated in the Linden route using NDT map [3] for localization and an Autoware [4] path following algorithm with pure pursuit steering to follow the AV shuttle path. The idea is that users will be able to easily implement and test their own AV functions in this environment. How the same approach can be extended to HiL and AViL simulations is presented in the paper along with how to add traffic, disturbances, uncertainties and faults to the overall system including the map for testing purposes.

Research on developing AV simulators or adapting available simulators to handle AVs also is a current topic of research with many recent papers and work being available in the literature. Some of these papers focus on synthetic data generation for use in training neural networks [5] while others focus on developing highly realistic models of the sensors being used, for example, PreScan. Other simulators focus on vehicle dynamics aspects of AVs and the effect of different road conditions such as CarSim. Also, some research work have been done upon ITS simulation, such as [6] [7]. While all of these approaches and the many commercially available AV simulation tools are very useful, they are not pursued here as our aim is to build a unified simulation evaluation approach for AVs using open source and freely available tools. The motivation is that the same user communities that contributed to and made these freely available or open source tools reach widespread acceptance will also contribute to this open source and hopefully user supported and sustained approach to building simulation tools and methods for AV evaluation. It should also be noted that the AV simulation approach based on open source or freely available tools presented here can also easily be incorporated into commercially available tools if desired.

The organization of the rest of this paper is as follows. In the next section, two alternative pipelines of three dimensional modeling are presented for creation of a simulation environment that realistically describes the real world environment information using available game engines. In the following section, an explanation of how to use the simulation environment for AV and AV shuttle pre-deployment testing is presented for the specific example of the Linden Residential Area and related results are shown. In the third section, simulation using Nvidia Drive PX2 board and the DriveWorks Library is implemented and results are shown. The paper ends with conclusions being given in the last section.

## Three Dimensional Environment Modeling

Three dimensional models represent the shape of an object that is rendered with a meshed texture representing its realistic look. For research purposes, a 3D model does not only provide a precise visualization of the object but also helps the design or configuration of certain physical objects in a simulation environment. In the area of computer vision, a lot of research is being done on modeling object and items based on different sources information, such as camera image [8], or point cloud data based on lidar detection [9].

Apart from modeling a single object for different uses, in the area of autonomous vehicles research, an accurate 3D modeling of the target deployment area is essential for developing a simulation environment as it should have a comprehensive representation of the real world so that the perceived environmental information is closely emulating the real world. For MiL and HiL simulation and testing of autonomous driving functions and safety evaluation of a planned deployment using simulation tools, a realistic virtual environment is extremely important from many aspects, including use of a correct map of roads, positions of objects in the environment, road infrastructure and features in the environment and the ability to add traffic and Vulnerable Road Users (VRU) like pedestrians and bicyclists. So far, various modeling algorithms have been presented for urban environment building based on satellite images or point cloud data [10]. Many companies have been working on modeling the real world in 3D for commercial use with decent effect and visualization for users, such as the Google map [11, 12]. High definition maps that are specifically built for autonomous driving are also available for highways from companies like HERE [13] and TomTom [14]. However, those maps are not publicly accessible so that it is inconvenient to extract a 3D urban model or a highway model from them. It is also costly to run those 3D modeling algorithms regarding computation resources and time.

In this section, two different ways of creating a 3D model of the real world for simulation are presented. Instead of using the commercially available tools mentioned above, we utilized the freely available OpenStreetMap (OSM) as our map source to create the 3D models for real world along with different mesh generation and rendering algorithm to build an accurate 3D model for testing and validating AV and AV shuttle deployment on a designated route.

### *Three Dimensional Environment Modeling Based on the OpenStreetMap*

The OpenStreetMap [15] is a publicly accessible map, all the users have access to the geodata underlying the map and can easily export the map as paper maps and electronic maps containing road network information, building information and road feature information that can be used for geocoding of address and place names as well as for route planning. The road network in the OSM map is relatively complete so that it can be used in navigation and traffic simulations. However, there are also some disadvantages of using OSM. Since OSM is an open source map and based on user contributions, the map lacks maintenance, therefore, there are problems that need to be fixed. The problems include incorrect road information and missing building information in some residential areas such as the Linden Residential Area region in Columbus. Thus, before modeling the simulation environment, the OSM map has to be updated and shared with the user community. In our work, we added the missing map information along the route in the Linden Residential Area and updated the map in the server, and we'll keep updating maps in OSM for the AV shuttle deployment sites in the city of Columbus.

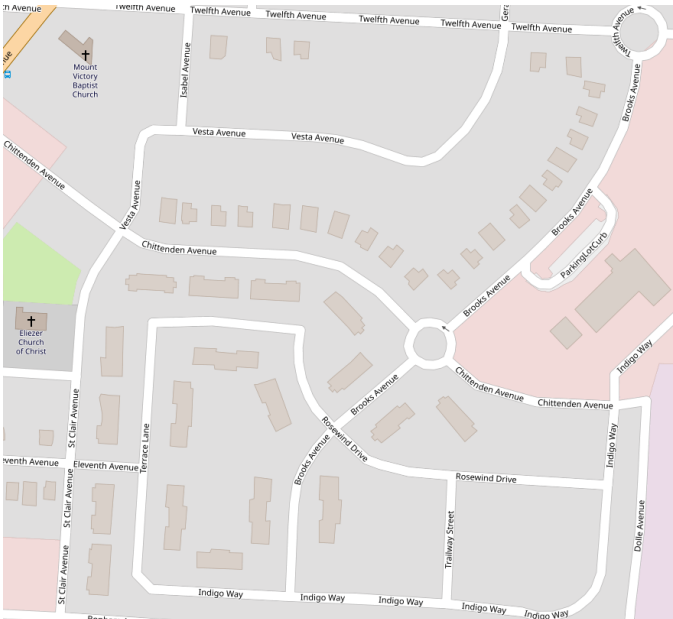

Figure 1. Bird view of part of the AV shuttle route in Columbus from OSM.

An example of the OSM exported file is shown in Figure 1 which contains the road network, terrain information and buildings information. It should be noted that the gray areas in Figure 1 are where buildings were added by the authors to update the OSM map. In general, an update and modification of the OSM map may have to be done before using it for building a simulation environment.

The 3D modeling is implemented in the Unity engine first [16]. The Unity engine is a cross-platform game engine developed by Unity Technologies for real-time, three dimensional graphics visualization. The engine has been adopted in many aspects outside the field of video games such as filming, automotive CAD, architecture and construction. With the rise of the autonomous vehicle industry, the Unity engine is now applied more and more in the field of autonomous driving. The Unity engine is one of the platforms that can be used for visualization of the autonomous driving environment and for extracting autonomous vehicle raw data. As shown in Figure 2, the raw data of OSM in the interest area is realized in the Unity engine but the first effects and visualisation are poor to work with. Thus, in this paper, a pipeline for 3D model generation and rendering is presented for a better simulation environment and synthetic sensor data creation as compared to the direct use of the raw OSM data.

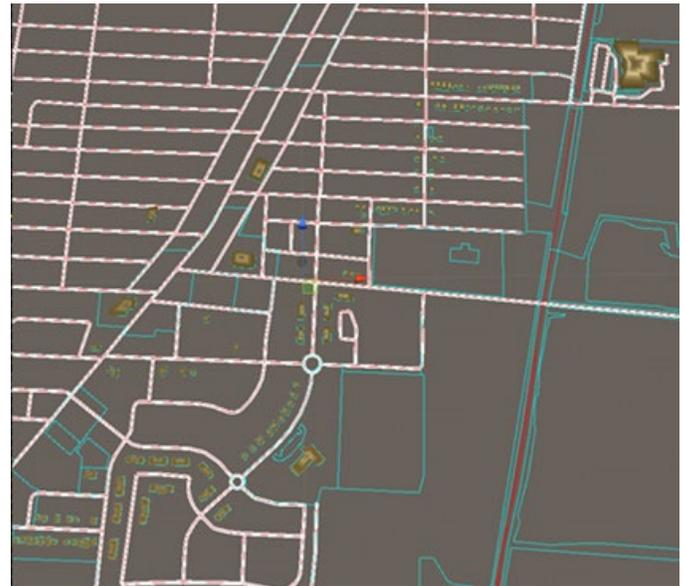

Figure 2. Visualization of raw OSM data in Unity engine of the Linden Residential Area AV shuttle route in Columbus.

For 3D modeling of real world data, different steps are needed starting from importing real world map data and ending in mesh rendering as shown in Figure 3. The procedure in Figure 3 is used here to create a 3D model of real world in the geo-fenced area of interest for AV shuttle simulation testing. This procedure is explained in more detail next.

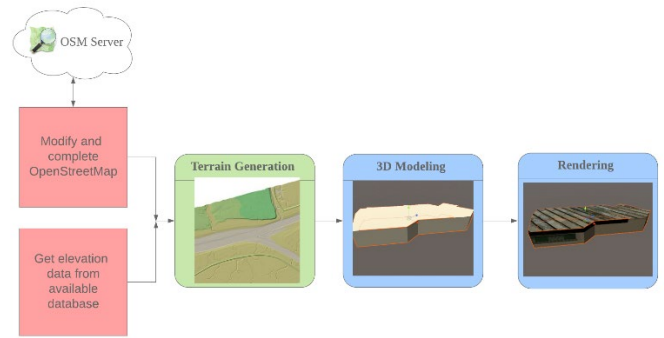

Figure 3. Pipeline for 3D modeling of the area of interest

1. OSM edition: Edit and update the OSM map data. As described in the previous part, before using the OSM map data, the users have to update and check the OSM to fix the problems in the OSM data. Then, upload the updated map into the OSM server so that it is available to the user community.

2. Elevation information: Get the elevation information from an available elevation dataset to generate a terrain with complete information. There are many publicly available datasets such as the SRTM [17]. The OSM map together with elevation map can help with terrain generation with actual road grade information for AV simulation.

3. Modeling: 3D modeling of the environment and the objects that are on the chosen terrain. At this step, the road network, objects and road features are taken care of. The 3D model generation is based on the OSM data regarding buildings and infrastructure. The 3D model of an object is made up using multiple surfaces describing the manifold of an object in the real world.

4. Texturing: Rendering the 3D model in the simulation environment. Texture and physical property of objects are attached to the 3D model created in the previous step in the geo-fenced area of interest. Based on real world's sensor data including images and point cloud, the textures have similar pixel-wise features in the synthetic sensor data when compared with sensor data derived from real world perception.

A complete 3D model including road network, environment features, buildings and traffic infrastructure was created based on OpenStreetMap data as shown in Figure 4 for the Linden Residential Area AV shuttle route of Columbus. Various traffic scenes and different areas of interests can be imported and added into simulations to emulate the real world traffic scenarios. The environment can be exported as a compactible filmbox (*.fbx) file and utilized in gaming simulators including Unreal and Unity Engines.

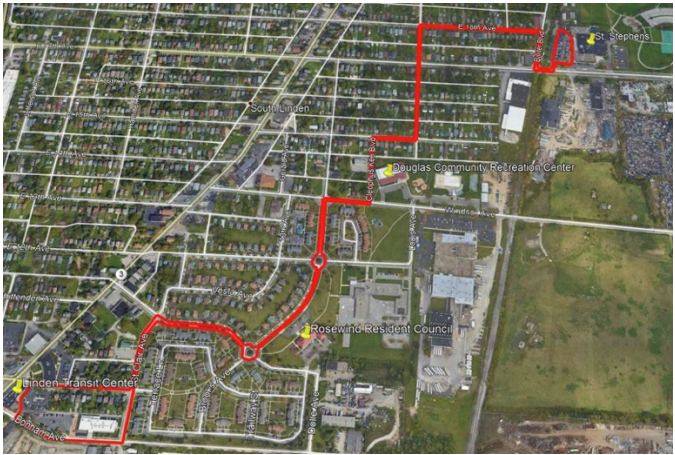
(a)

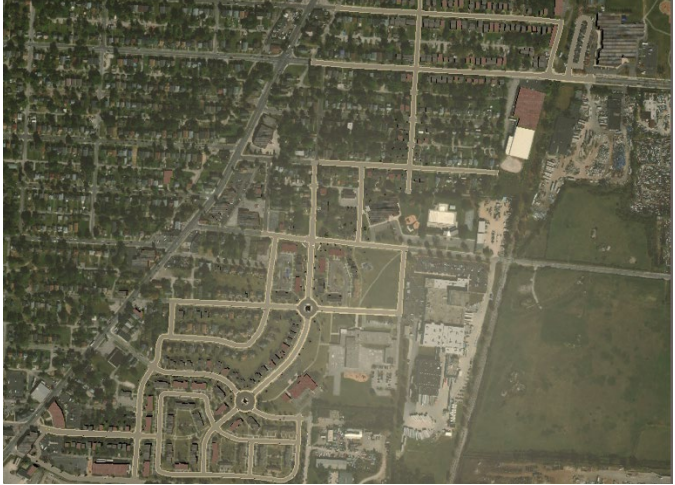
(b)

Figure 4. (a) The real world environment; (b) The 3D model simulation environment. Road surface and road features along the AV shuttle route are modeled realistically.

## *Three Dimensional Modeling Based on Point Cloud Rendering*

In this section, we will present another way of 3D modeling and briefly explain the methodology and an efficient workflow to create the 3D environment for AV simulations using existing state-of-art libraries and tools. The generated 3D model for the AV shuttle route part of the Linden Residential Area is shown in Figure 5. The overall aim is to use time-stamped lidar data to reconstruct the surface of the entire environment and then to add texture details from Google Maps (Google Earth Studio) and road details from OpenStreetMaps. One may also use custom collected images for realistic texture for simulation.

1. Pre-Processing: Reduce the redundant point cloud data using Poisson Sampling Disk [18] to boost the processing speed without considerable loss in reconstruction. If preferred, one may do this step after the mapping process. Nearly 40 - 45% reduction of our points is achieved at the end of this step. Optional step - One may also remove point clouds associated with other vehicles using SqeezeSeq [19] or other methods in order to eliminate them during the mapping process.

2. Mapping: Use the mapping algorithm to generate a single point cloud of the entire scene. We used NDT (Normal Distribution Transform) based mapping to generate the dense point cloud. One may repeat Step 1 again to reduce the point cloud for faster computation. In particular, we used sampling disk [20] and Voxel sampling methods.

3. Meshing: Generate the mesh from the point cloud. First, calculate appropriate vertex normals heuristically based on lidar resolution, size of the environment and other such factors. Generate the mesh using the surface constructed with the generated normals. One may have to calculate appropriate normals for better results. In our case, Ball-Pivoting surface reconstruction [21] gave the expected results.

4. Cleaning and Smoothening: The mesh generated in the previous step might end up with voluminous faces. We used Quadratic Edge Decimation [22] to reduce the triangles without much loss to the reconstruction. In addition, one might have to remove duplicate edges and single point triangles. We used the Laplacian filter to smooth out the surface of the mesh.

We collected satellite images of our route using Google Earth Studio. One can also collect their own data using drones or vehicles with cameras in order to construct a more realistic and recent replication of the environment. If images are taken from Google Earth, one might not be able to recover certain details in regions such as those under the trees as they will be occluded in the input images themselves. For this reason, we recommend one to collect their own data if possible. In our case, however, we used Google Earth Studio for data collection. In particular, we collected nearly four thousand images of our route.

In the upcoming steps, we will discuss only the important tasks that were used to achieve the desired results, while skipping the intermediate steps like feature extraction, point-to-point correspondence and feature matching.

5. Structure From Motions (SfM) and Depth Map: Generate dense point cloud of the scene using the SfM [23, 24] approach. We used Meshroom's implementation of SfM for its ability to handle large data, GPU enabled libraries and interface. One might also use VisualSFM, if preferred. It is also possible to reconstruct the environment using this dense point cloud, however, such reconstructed environment will inherit the above-mentioned drawbacks including reconstruction loss due to occlusion. For this reason, we prefer the mesh generated from lidar data for its measurement accuracy.

6. Texturing: In this final step, we add the texture (SfM) to the lidar based mesh using Least Square Conformal Maps [25]. We used Blender to fix the scale and origin. The generated mesh can then be exported as a compactible filmbox (.fbx) file and utilized in gaming simulators including Unreal and Unity Engines.

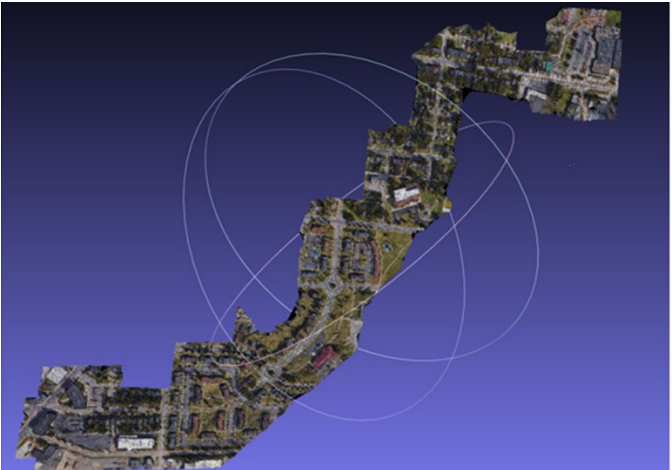

Figure 5. 3D model of the AV shuttle route rendering from lidar point cloud

## Simulation Development for AV deployment

The 3D modeling of the geo-fenced area containing the routes for AV shuttle deployment was presented in the previous section. Testing and validation using this virtual simulation environment is presented in this section, to explain how the simulation environment can be used for AV shuttle pre-deployment testing

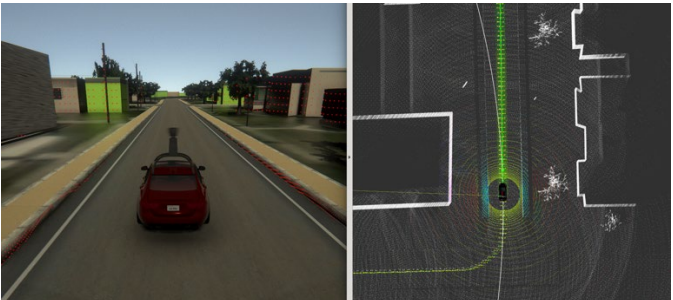

Figure 6. The simulation scene on the Linden AV shuttle route

Vehicle simulation tools have been well developed for many years for high fidelity vehicle dynamics evaluation, for example, Carsim. However, with the rise of research in autonomous vehicles and automated driving functionality, traditional vehicle simulators hve started to lack in the capability of simulating sensor suites for autonomous vehicles and the flexibility of generating various traffic scenarios which make it difficult to use them for validation and testing of automated driving scenarios. To solve this problem, open source intelligent transportation simulation tools based on game developing platforms have been developed recently for MiL, HiL and AViL testing before AV deployment on different routes. Both the Unity engine and the Unreal engine environments were used for the modeling which can be used in the freely available simulators LG SVL [26] and CARLA [27]. The fundamental autonomous driving functions including localization of the ego autonomous vehicle and path following along the defined AV shuttle route for pre-deployment tests are studied next.

## Nvidia Based HiL Simulation

In this section, the HiL Simulation using Nvidia Drive PX2 driving hardware with the simulator and 3D model of real world data is presented. The DriveWorks library with the Nvidia Drive PX2 is capable of running deep neural networks for perception in the AV shuttle deployment. In our simulation, synthetic images from the virtual camera in unity simulation environment are fed to the Drive PX2 computer to be processed using various algorithms for free space detection, lane detection and object detections.

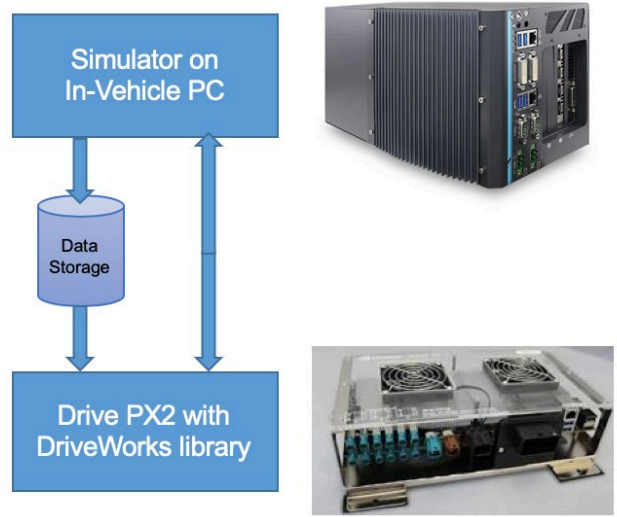

Figure 7. The diagram of Simulation using Nvidia Drive PX2

The block diagram of simulation with Drive PX2 is shown in Figure 7. The data is fed to the Drive PX2 from the in-vehicle PC with simulators built-in. For now, the simulation is done with unity based simulation to validate the simulation environment.

Simulation results from the Nvidia Drive PX2 are shown in the following figures. The simulation results shown here illustrate that the simulation environment for AV function test is realistic and ready for future use. Using synthetic data from the simulation environment of the Linden Residential Area and other environment, simulation is done and the result of free space detection, lane detection, traffic light detection, object detection and classification are shown in the results. In Figure 8, the free space is shown as the result of free space detection. With the border of green and red, it provides information about the space that is available for the AV shuttle to move in. The lane detection simulation result in Figure 9 shows the driving lane between red and green road boundaries. The object detection shown in Figure 10 and traffic light detection in Figure 11 show results upon object detection and classification with the DriveWorks library. As for object detection and traffic light detection, a bounding box will show up with different colors with respect to different class of objects. Totally, five classes of objects are recognized including car, bicycle, traffic sign, pedestrian and traffic lights. All these detections can run at the same time in real time and can be used by the autonomous driving algorithm part of an autonomous shuttle.

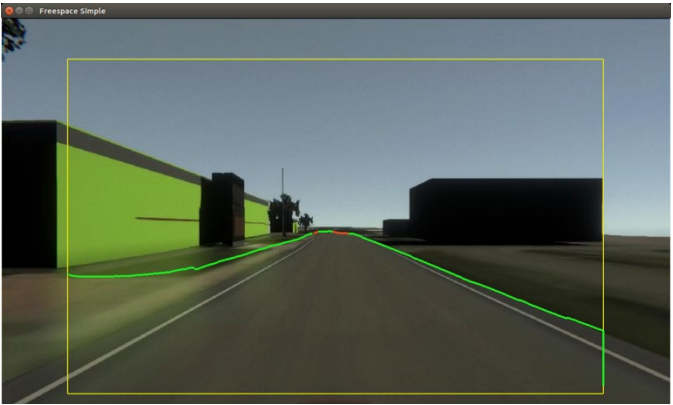

Figure 8. Free space detection using Drive PX2 in Linden area

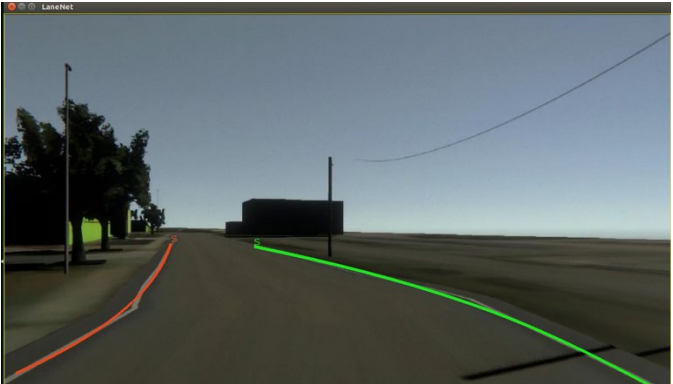

Figure 9. Lane Detection using Drive PX2 in Linden area

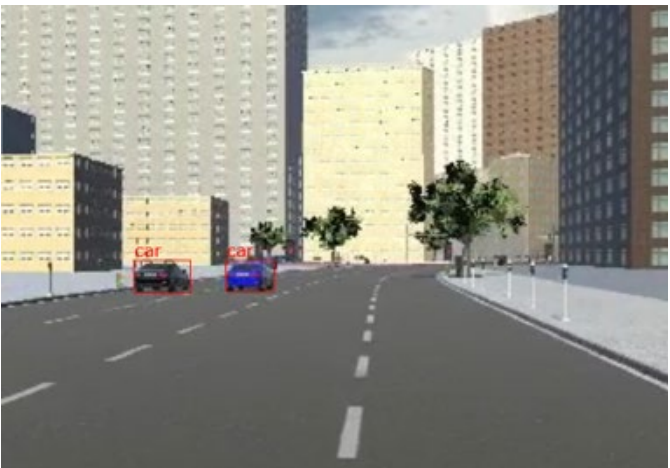

Figure 10. Object Detection using Drive PX2

## Simulation in Unity Engine

The Unity engine based simulation environment developed here for AV simulation testing uses the freely available and open source LG SVL simulator. In the simulator, a complete sensor suite for autonomous vehicles including three dimensional (3D) lidar, cameras, radar, GPS and IMU are provided. The set of synthetic data perceived in the simulation environment from 3D modeling is used for automated driving of the AVs in the simulator.

Here, the Unity based MiL simulation is an AV simulation that collaborates with the freely available and open source Robot Operating System (ROS) [28] for localization and vehicle control functions. The communication between ROS and Unity engines utilizes the ROS bridge function as shown in Figure 12.

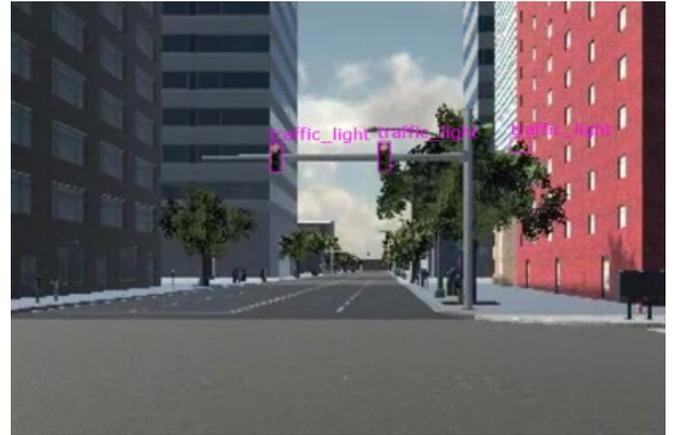

Figure 11. Traffic light detection using Drive PX2

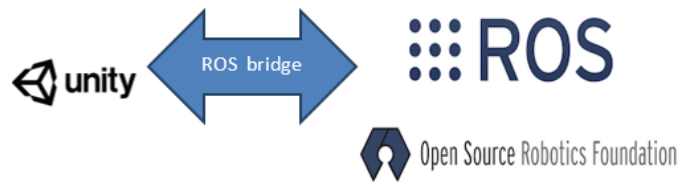

Figure 12. Diagram of Unity engine based simulation

Here, in this simulation, the localization of the ego vehicle and basic path following control algorithm are implemented for AV shuttle pre-deployment testing and results are shown for explaining the capabilities of simulated AV driving.

**Map Matching Based Localization**

Localization of the ego AV is the fundamental AV driving function in an autonomous driving task as it provides current location and pose of the vehicle. Various localization algorithms have been developed including GPS unit based global localization and lidar or camera based SLAM algorithms [29] to derive location information within the sensors' field of view (FOV). In order to get precise global location using the GPS unit, high accuracy GPS rather than the cheap on board GPS units available in current vehicles is required. As high accuracy GPS units are currently at a very high price, such method of localization makes it costly to build an AV or AV shuttle. GPS also does not work in areas of poor GPS satellite connection. On the other hand, the SLAM method can only provide local map and location information and sometimes can be computational expensive. Therefore, there are also challenges of implementing the SLAM algorithm when the global information is needed and computational power on the AVs is limited.

Hence, map matching based localization algorithms are researched and developed to provide a local position information which can be used to infer global location. The map matching algorithm does not require strong computational power as map building is an a priori and hence offline process which saves the computing resource from this tedious task during AV operation.

In the simulation of AV shuttle pre-deployment tests, a map matching localization algorithm based on Normal Distribution Transformation [3] using 3D point cloud map generated based on NDT mapping algorithm in the freely available Autoware [4] tool is used here. Then, the map is utilized for localization in the map reference frame of the 3D map for local information which will be used for global coordinate generation.

*Normal Distribution Transformation based Scan Matching*

The normal distribution transformation (NDT) is a way of representing point cloud as a probability density function in order to match different frames of point cloud. As shown in Figure 13, the normal distribution is used for representing the points in a cell that can describe the density and features within that cell. Then, we find the optimal transformation between the current point cloud and the target point cloud.

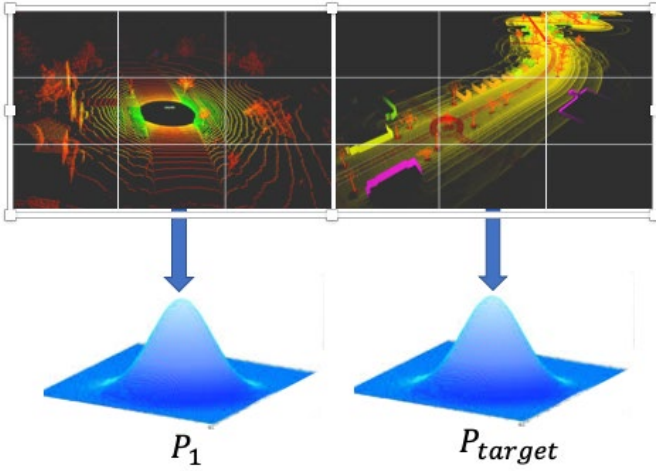

Figure 13. Sketch of NDT towards different point cloud

For a cell in the point cloud, the following equations are used for a transformation from a 3D point cloud to probability density function with a multi-variable distribution function having a set of measure variable $\vec{x}_{k=1,2,3,\dots,n}$

$$p(\vec{x}_k) = \frac{1}{(2\pi)^{D/2}\sqrt{|\Sigma|}} \exp\left(-\frac{(\vec{x}_k - \vec{\mu})^T \Sigma^{-1}(\vec{x}_k - \vec{\mu})}{2}\right)$$

(1)

$$\vec{\mu} = \frac{1}{m}\sum_{k=1}^{m} \vec{y}_k$$

(2)

$$\Sigma = \frac{1}{m-1}\sum_{k=1}^{m}(\vec{y}_k - \vec{\mu})(\vec{y}_k - \vec{\mu})^T$$

(3)

In the equation above, $\vec{y}_{k=1,2,3,\dots,m}$ are the positions of the point cloud contained in the cell. As the NDT is used for scan matching between the current point cloud and target point cloud, a score function regarding rotation and translation of the current point cloud is to be optimized as shown in equation (4),

$$s(\vec{p}) = \sum_{k=1}^{n} p(T(\vec{p},\vec{x}_k))$$

(4)

which the pose $\vec{p}$ is the tuple of rotation and translation from current scan to the reference scan, carried out by the transformation function $T(\cdot)$. Optimizing the score function, a scan matching for finding the transformation from current point cloud to reference point cloud is obtained by NDT. For a 3D transformation function, the Euler orientation sequence $z - y - x$ is applied for optimization. The NDT scan matching approach is illustrated in Figure 14.

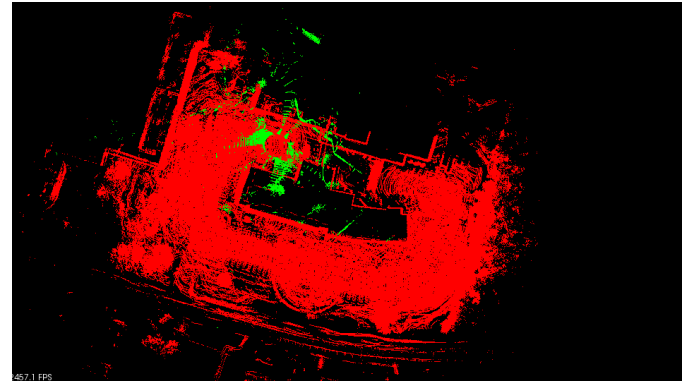

Figure 14. An example of scan matching. The current point cloud (green) is matched to the reference point cloud (red)

*3D Mapping and Localization*

Different frames of 3D point cloud perceived by the lidar sensor are matched and overlapped for generating the 3D point cloud high definition map using the Autoware platform as it is an open source solution. Using the synthetic data in the Unity Engine based simulator, the 3D map contains important road features for localizing the vehicle and emulates the real world environment. The mapping results utilizing collected lidar data with the AV sensor suite and synthetic data are shown in Figure 15. The map based on synthetic data filters redundant information in the collected lidar data and it keeps useful road features for localization use.

As for localizing the ego vehicle, the normal distribution transformation based localization is utilized here. The vehicle localization follows the flowchart in Figure 11 and is generated based on the transformation between the current point cloud and the map. Given the pose of the lidar sensor on the vehicle $T_{lidar}$, the vehicle location containing rotation and translation from the origin of the map frame is

$$T(rot, t) = T_{lidar} * T_{ndt}$$

(5)

If the GPS coordinate $T_{o\_gps}$ is known based on the measurement of cheap on board GPS unit, the global coordinate is derived by equation (6)

$$T_{global} = T_{o\_gps} * T_{map} \tag{6}$$

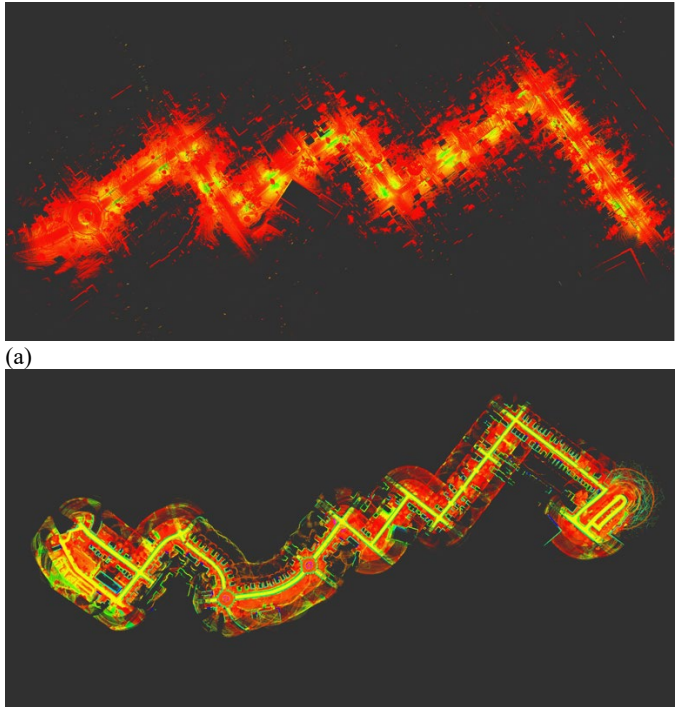

(a)

(b)

Figure 15. (a) 3D point cloud map generated by Lidar data collected in real world; (b) 3D point cloud map in the simulator by synthetic data

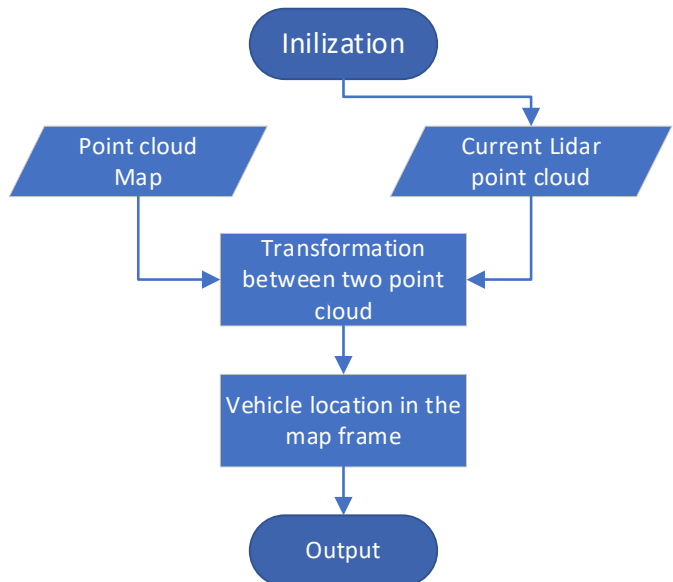

Figure 16. Flow chart of map matching based localization

**Path Following**

A pure pursuit [30] path following algorithm is deployed for the simulated autonomous driving along the pre-defined AV shuttle route that can be easily replaced by the user's own algorithm. It uses the single track vehicle model and steers the ego AV shuttle towards the nearest feasible way points on the route.

Knowing the current location of the vehicle, it will find the closest path points and the goal point of the look ahead distance. With the two way points, the curvature is computed for the vehicle to steer the vehicle using the steering angle given by

$$\delta = \tan^{-1}\left(\frac{2L\sin(e(t))}{l_d}\right) \tag{7}$$

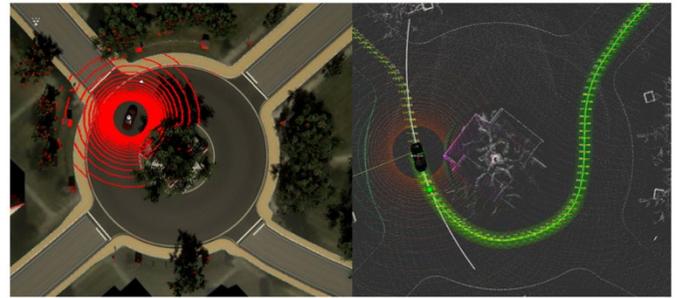

Figure 17. Pure pursuit path following in the simulation at a roundabout. The green point is goal point of the vehicle while moving along the closet way points

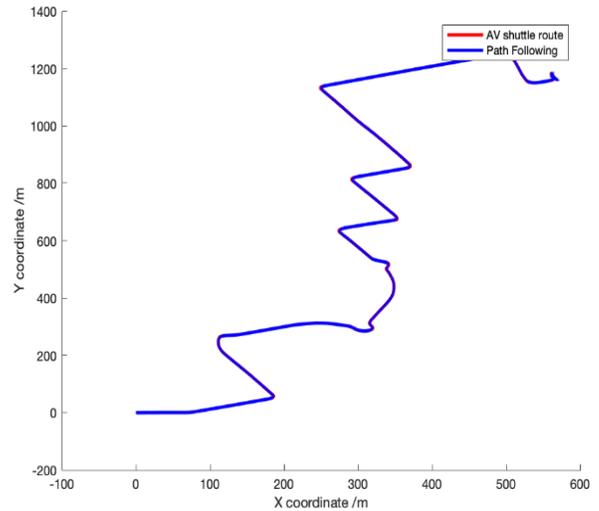

Figure 18. AV shuttle route comparing with pure pursuit algorithm path following

In equation (7), $L$ is the length of the vehicle bicycle model in between its two axles, $e(t)$ is the angle between heading and the closet way point, $l_d$ is the look ahead distance of the pure pursuit algorithm. $6\ m$ of look ahead distance was used in the simulation test. The steering command is generated and sent to the ego AV in the simulator. The vehicle is set to move at cruise control mode in order to move at constant speed. As it is simulating the AV shuttle service, the speed is up to 25 miles per hour. Figure 17 shows the simulation scenarios of path following in autonomous driving and Figure 18 shows a comparison of the AV shuttle route in Linden with the NDT localization based path following using simple pure pursuit algorithm at the speed of 25 miles per hour. As the two paths shows in the figure, the error between the reference path (red) and path following path (blue).

## Simulation in Unreal Engine

Similar autonomous driving functions including localizing and path following are applied in the CARLA simulator for the AV shuttle pre-deployment test in the Unreal engine. The freely available and open source CARLA simulator is used with the Unreal Engine platform. CARLA has a powerful Application Programming Interface (API) that allows users to control all aspects related to the simulation including traffic generation, pedestrian behavior, weather conditions and sensors. Autoware routines including Path Planner and Vector Map Navigation were used in the Unreal platform CARLA AV driving simulations are illustrated in Figure 19.

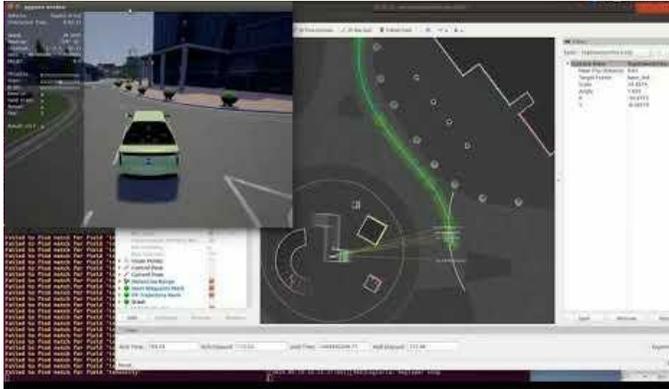

Figure 19. A screen shot of the Unreal engine based AV simulation in CARLA

## Conclusions

In this paper, the simulation environment development for AV or AV shuttle pre-deployment testing of low speed autonomous driving is presented. The way of modeling the real world map data is flexible and adjustable for creating various traffic scenarios. It provides a solution for validating and testing AV driving functions with some simulation results shown here. As it is easy to modify, different algorithms including planning and perception algorithms can also be tested in such simulators. This AV simulation environment is currently being extended by adding user defined and adjustable disturbances and unexpected events to test the robustness and resilience of autonomous driving software.

The work presented here was based on using open source and publicly available simulators using freely available game engines for visualization. While this is a lower cost method as opposed to using commercially available software and maps, there is a trade-off between cost and research staff resources used as this reduced cost approach requires more development time. The pre-deployment evaluation methods introduced in this paper can be incorporated and used in developing, simulating and evaluating methods as diverse as controls, automotive controls including powertrain control, advanced driver assistance systems, connected vehicles and autonomous driving systems [31-85] in extensions of this work.

## Acknowledgments

The authors would like to thank the Smart Campus organization of the Ohio State University and Smart Columbus for partial support of the work presented here. The author also want to thank Nvidia for donating two Nvidia Drive PX2 device for our research.

## Definitions/Abbreviations

| | |
|---|---|
| **AV** | Autonomous Vehicle |
| **MiL** | Model in the Loop |
| **HiL** | Hardware in the Loop |
| **AViL** | Autonomous Vehicle in the Loop |
| **SAE** | Society of Automotive Engineers |
| **3D** | Three dimensional |
| **GPS** | Global Positioning System |
| **IMU** | Inertial Measurement Unit |
| **Lidar** | Light Detection and Ranging |
| **SLAM** | Simultaneous Localization and Mapping |
| **NDT** | Normal Distribution Transformation |
| **FoV** | Field of View |
| **ROS** | Robot Operating System |
| **API** | Application Programming Interfce |